\documentclass[conference]{IEEEtran}
\IEEEoverridecommandlockouts

\usepackage{url}
\usepackage{hyperref}
\usepackage{graphicx}
\usepackage{booktabs}
\usepackage{listings}
\usepackage{color}

\definecolor{dkgreen}{rgb}{0,0.6,0}
\definecolor{gray}{rgb}{0.5,0.5,0.5}
\definecolor{mauve}{rgb}{0.58,0,0.82}

\lstdefinestyle{customc}{
  belowcaptionskip=1\baselineskip,
  breaklines=true,
  frame=L,
  xleftmargin=\parindent,
  language=Python,
  showstringspaces=false,
  basicstyle=\footnotesize\ttfamily,
}

\usepackage{multirow}
\usepackage[table,xcdraw]{xcolor}

\usepackage{cite}
\usepackage{amsmath,amssymb,amsfonts}
\usepackage{algorithmic}
\usepackage{graphicx}
\usepackage{textcomp}
\def\BibTeX{{\rm B\kern-.05em{\sc i\kern-.025em b}\kern-.08em
		T\kern-.1667em\lower.7ex\hbox{E}\kern-.125emX}}

\begin{document}

\title{Extending Variability-Aware Model Selection with Bias Detection in Machine Learning Projects}
\date{June 2022}
\author{\IEEEauthorblockN{Cristina Tavares, Nathalia Nascimento, Paulo Alencar, Donald Cowan} 
\IEEEauthorblockA{\textit{David R. Cheriton School of Computer Science} \\
\textit{University of Waterloo}\\
Waterloo, Canada \\
\{cristina.tavares, nmoraesdonascimento, palencar, dcowan\}@uwaterloo.ca}}

\IEEEoverridecommandlockouts
\IEEEpubid{\makebox[\columnwidth]{979-8-3503-2445-7/23/\$31.00~\copyright2023 IEEE \hfill} \hspace{\columnsep}\makebox[\columnwidth]{ }}

\maketitle

\IEEEpubidadjcol

\begin{abstract}
Data science projects often involve various machine learning (ML) methods that depend on data, code, and models.
One of the key activities in these projects is the selection of a model or algorithm that is appropriate for the data analysis at hand. ML model selection depends on several factors, which include data-related attributes such as sample size, functional requirements such as the prediction algorithm type, and non-functional requirements such as performance and bias.
However, the factors that influence such selection are often not well understood and explicitly represented. This paper describes ongoing work on extending an adaptive variability-aware model selection method with bias detection in ML projects. The method involves: (i) modeling the variability of the factors that affect model selection using feature models based on heuristics proposed in the literature; (ii) instantiating our variability model with added features related to bias (e.g., bias-related metrics); and (iii) conducting experiments that illustrate the method in a specific case study to illustrate our approach based on a heart failure prediction project. 
The proposed approach aims to advance the state of the art by making explicit factors that influence model selection, particularly those related to bias, as well as their interactions. The provided representations can transform   model selection in ML projects into a non ad hoc, adaptive, and explainable process.

\end{abstract}


\section{Introduction}

Data science projects increasingly rely on machine learning approaches, and these projects often involve various machine learning (ML) methods that depend on data, code, and models \cite{mehrabi2021survey}.
One of the key activities in these projects is the selection of a model or algorithm that is appropriate for the data analysis at hand. ML model selection varies based on several factors, which include data-related attributes such as sample size, functional requirements such as the prediction algorithm type, and non-functional requirements such as performance and bias.

However, the factors that influence such selection are often not well understood and explicitly represented, particularly
in terms of bias detection \cite{fu2020ai,pessach2022review, pagano2022bias}. ML bias can be caused, for example, by the lack of complete and accurate data, and comprise several forms, including racial or gender bias and age discrimination, but can also be related to factors that go beyond technology itself and include human aspects such as trust \cite{fu2020ai,fahse2021managing, adchariyavivit2021tools}. 

This paper describes ongoing work on extending an adaptive variability-aware modeling method with bias detection in ML projects.
The method involves: (i) modeling the variability of the factors that affect model selection using feature models based on heuristics proposed in the literature; (ii) instantiating our variability model with features related to bias (e.g., bias-related metrics); and (iii) conducting experiments that illustrate the method in a specific case study to illustrate our approach based on a heart failure prediction project. 

In this context, two key research questions become relevant: (RQ1) How to extend variability-aware model selection with added features related to bias? (RQ2) How instantiate the model and conduct experiments involving bias-related features in a specific case study involving heart failure prediction?

The variability-aware method captures interactions among various abstractions that impact algorithm selection (e.g., dataset, prediction type, and its outcomes such as performance and bias). For example, the data sample size may increase and result in the selection of different algorithms. Further, the prediction category may change as additional quantitative data become available, new important features are discovered and can be added to the dataset, or changes in outcomes (e.g., accuracy) may result in changes in the models (e.g., the drift problem \cite{lu2018learning}. In more dynamic settings, according to Hummer et al. \cite{hummer2019modelops}, while in the classical application lifecycle, new builds are triggered by code base changes, in the AI application lifecycle, new builds could be triggered by data or code changes, which may activate a retraining process or require replacement of the current ML model for a new one.

The proposed method aims to advance the state-of-the-art by making explicit factors that influence model selection, particularly those related to bias, as well as their interactions. 
In general, the approach can lead to an improved understanding of the factors that influence the model selection, particularly when it comes to bias, how these factors explicitly affect the selection, and how adaptive factors can be represented and automated. This improved understanding can result in a project model selection process that is less implicit and make its application more productive.
The approach also advances the state-of-the-art by introducing an adaptive and explainable process. Introducing adaptive processes provides support for dealing with the variations that occur in model selection. In addition, introducing an explainable process provides support for accountability, making the reasons why a method has been chosen. Finally, the proposed adaptive method can ultimately constitute a foundation for the creation of novel dynamic software product lines to support the model selection process.

The paper is structured as follows. Section II describes the research background and related work. Section III presents our variability-aware ML model selection approach to assess bias in ML projects.  Section IV presents a case study that illustrates the applicability of the approach. Finally, Section V presents conclusions and future work.

\color{red}
\color{black}

\section{Background and Related Work}

\subsection{ML Model Selection}
Modeling is one of the phases of the data analysis process, which consists in selecting and applying several modeling techniques and their algorithms to solve a problem until specific quality criteria are satisfied \cite{chapman2000crisp}. Among the multitude of types and algorithms of model techniques, ML-based data analysis is a widely adopted paradigm that has been applied in data science problems in several domains, including health, business, and smart cities \cite{francca2021overview}. Considering that different algorithms based on ML can be selected based on factors such as sample size, method category, and data types, selecting the appropriate algorithm is one of the most challenging steps in the context of the data model selection process. According to \cite{yao2018taking}, every aspect of ML-based analysis applications must be configured, indicating a need for new approaches and systems to automate the various phases of the data analysis process. 

Some heuristics have been proposed to capture criteria for choosing particular machine learning methods. These heuristics can help automate this selection if they are appropriately captured. Although heuristics have been provided to guide the ML data analysis modeling process phases, such as the algorithm selection phase, these heuristics have not been used to capture the variability of this phase. From a software engineering perspective, designing and implementing a software approach to accommodate possible variations in factors that affect ML algorithm selection based on heuristics can lead to a configurable solution.

\subsection{Variabiliy, Feature Models and Adaptation}
According to \cite{apel2013basic}, variability is defined as the ability to derive different products from a common set of artifacts. Variability-aware approach has been proposed in several areas, including software engineering, databases, and data warehouses \cite{valdezate2022ruva,Buhne2005,caplinskas2013variability, bouarar2015spl}. The foundation of these approaches rely on identifying and managing commonalities and variabilities that are abstractly captured through entities called features \cite{berger2015feature}. In the context of variability-aware approach applied to software engineering, a feature can be defined as distinct characteristics of a concept such as a system or a component, or either a logical component of
behavior that can be identified by a set of functional and non-functional
requirements” \cite{berger2015feature}. Thus, features are used in all stages of the software life cycle, supporting the software artifacts' reuse, variation, and structure \cite{apel2013development, apel2013basic}. The most popular form to model and represent variability in terms of common and variable attributes is the technique known as feature modeling \cite{Kang2013, apel2013development}. For the variability modeling analysis, the feature model is a widely used method to formally represent the description of features and their relations and constraints \cite{apel2013development}. 



   
 Therefore, feature models enable the representation of feature selection in a specific model instance, indicating these features can change based on the models both at design and runtime. Constraints in feature models can be seen as adaptation triggers in the sense that they denote conditions in which adaptations in variations occur.

Identifying and representing variabilities constitutes a basis for exploring opportunities for automation. 

\subsection{ML Bias Detection and Mitigaton}
ML prediction-based methods are increasingly being use by industry and governments to inform their decisions in several areas, including criminal risk prediction, heath care provisions, and mortgage lending \cite{mehrabi2021survey, stoyanovich2020responsible, pessach2022review, pagano2022bias}. The emergence of these techniques has led to concerns about bias involving these prediction techniques, that is, concerns about how fair they are in terms of social aspects such as race, gender, and class. In other words, there are concerns about the extent to which these methods may, in some cases, result in unfair discrimination or prejudice of some individuals or groups in favor of others \cite{pessach2022review, pagano2022bias}.

There are several possible sources of bias in ML methods. Some of these sources include data bias (e.g., historical or social), algorithm bias (e.g., popularity), and user interaction bias (e.g., behavioral). Several approaches have been propose to quantify bias in classification approaches, particulary those based on metrics \cite{pessach2022review, pagano2022bias}. Examples of such metrics include Equalized Odds, Equality of Opportunity, Demographic Parity, Individual and Differential Fairness, Disparate Impact, and K-Nearest Neighbors Consistency. Existing approaches can be used for bias detection and bias mitigation \cite{pessach2022review, pagano2022bias}.

\subsection{Related Work}
Related work encompasses topics such as automation, adaptation based on features, and bias detection and mitigation.
Regarding automation, the acronym MLOps (machine Learning operations) refers to an approach of designing and automating ML lifecycle processes. MLOps pipeline comprises a cycle of tasks such as data preparation, model creation, training, evaluation, deployment, and monitoring, which are grouped into three main basic procedures namely data manipulation, model creation, and deployment \cite{symeonidis2022mlops}. 

Regarding monitoring and reconfiguration, Martinez \cite{rivera_2010} proposes a model monitoring approach that considers the factors that can affect the performance of the model, such as a change in data context (e.g., culture, location, or time). The idea is to reconfigure the ML model based on the system performance dynamically. However, performance decrease is a critical problem for many applications data constantly changes over time, such as those related to healthcare scenarios, environmental monitoring, and air traffic control. 
Further, Nascimento et al. \cite{nascimento2021context} present an example of a reconfigured neural network based on the application context. The authors describe situations in which it is necessary to modify the neural network architecture itself (e.g., the number of layers) and not only to retrain it. 

In terms of bias, several works have explored the use of metrics as a basis for bias detection and mitigation \cite{pessach2022review, pagano2022bias}. Our approach aims to provide models that can adapt to features that influence ML model selection, and be explainable in the sense that the instantiation of the model clarifies which features are used in a specific context. Therefore, our model tackles bias in the context of ML model selection, making bias dependant on different types of explicitly represented of variabilities, including those related to the data, the algorithms, and the application functional and non-functional requirements.



\section{Extending a Variability-Aware ML Algorithm Selection Approach with Bias Detection} \label{Sec:Approach}
The proposed approach involves the following steps: (i) modeling the variability of the factors that affect model selection using feature models based on heuristics proposed in the literature; (ii) instantiating our variability model with added features related to bias (e.g., bias-related metrics); and (iii) conducting experiments that illustrate the method in a specific case study to illustrate our approach based on a heart failure prediction project. The steps (i) and (ii) of the approach will be described in the next subsections. 

\subsection{Heuristics for Model Selection}
\begin{figure*}[!ht]
	\centering
	\includegraphics[scale=0.77]{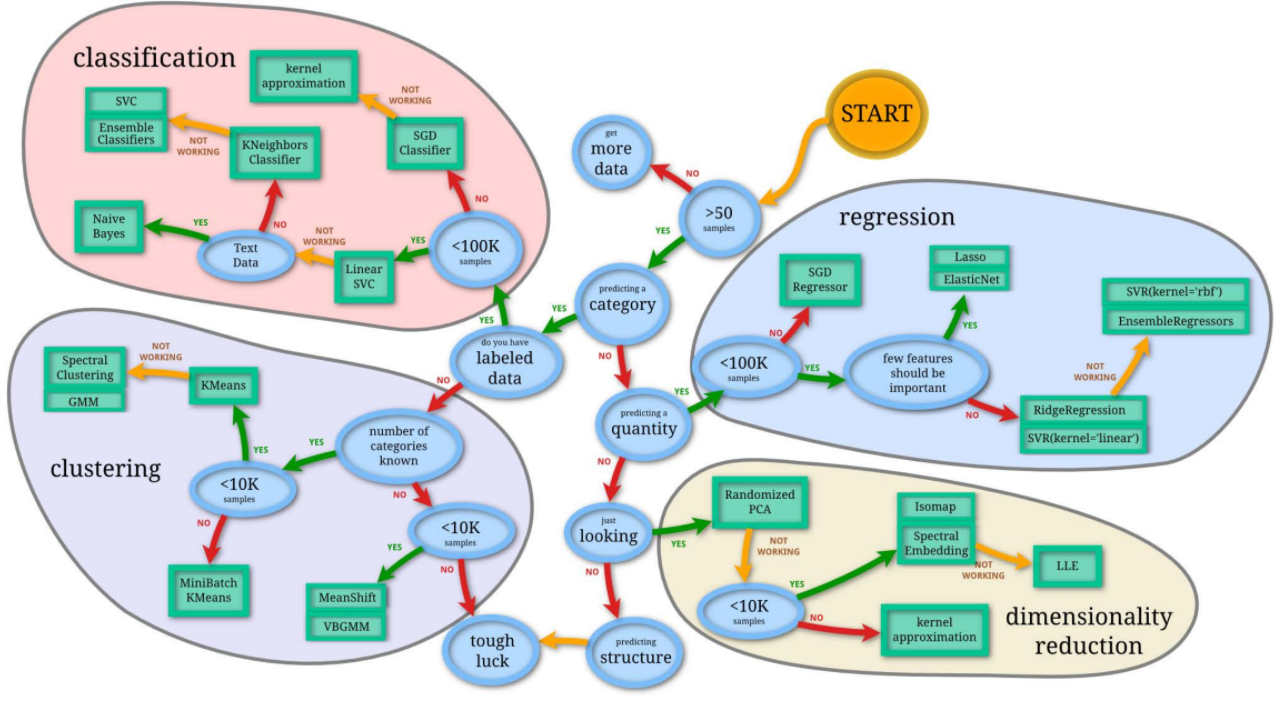}
	\centering
	\caption{Heuristics for selecting a machine learning algorithm from Scikit-Learn.}
	\label{fig:Scikit}
\end{figure*}

We focus specifically on heuristics provided in Scikit-Learn 1.0.1 \cite{scikit-learn}, which are shown in Figure \ref{fig:Scikit}. 

This flowchart provides heuristics to guide users on how to select specific algorithms. For example, if the sample size is equal to or greater than 50, the prediction is categorical, the data is labelled, and the sample size is less than 100,000, then a linear SVC algorithm is recommended. In case the SVC is not working, and the data is textual, the next recommended algorithm is the Naive Bayes.



\subsection{Feature Modeling using Feature Diagrams}
\begin{figure*}[!ht]
	\centering
	\includegraphics[scale=0.14]{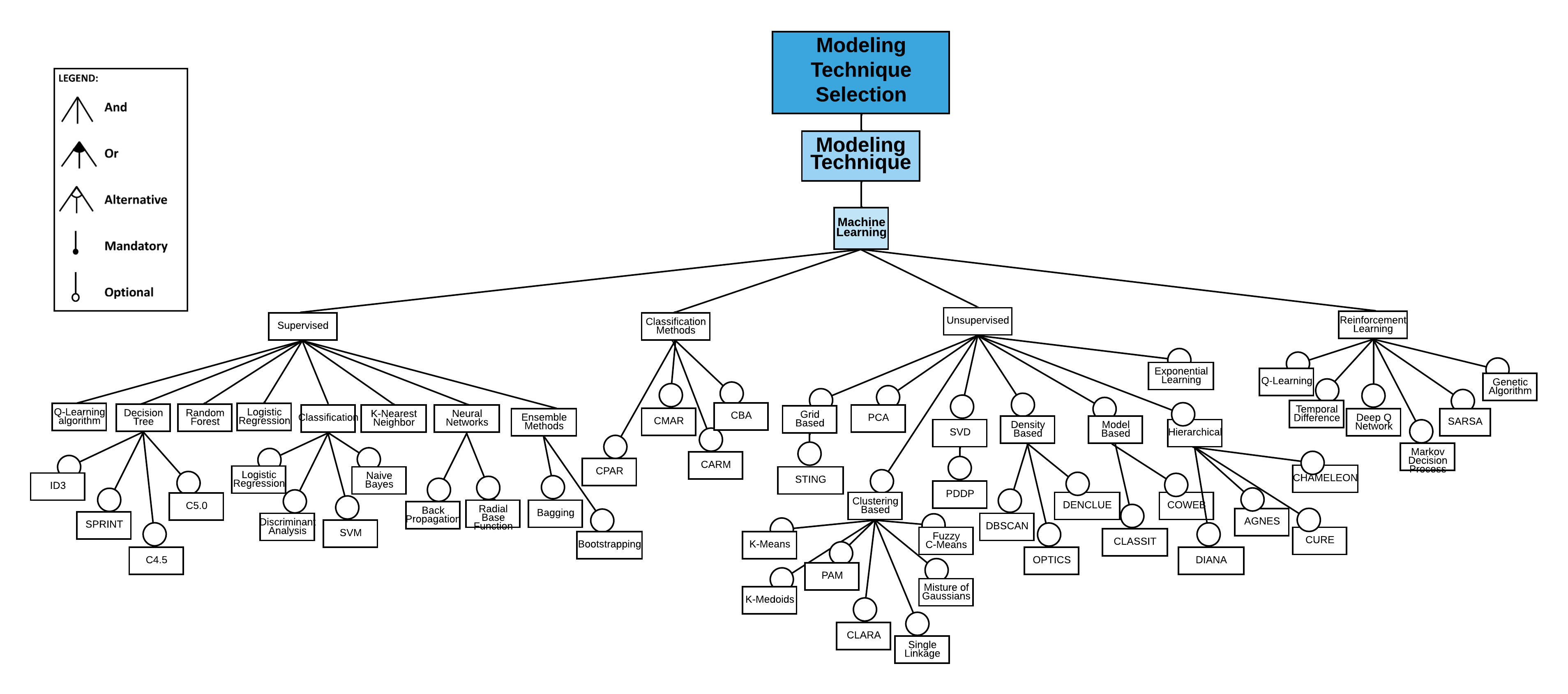}
	\centering
	\caption{Feature diagram for Modeling Technique Selection - Machine Learning.}
	\label{fig:ML Techniques}
\end{figure*}

\begin{figure*}[!ht]
	\centering
	\includegraphics[scale=0.24]{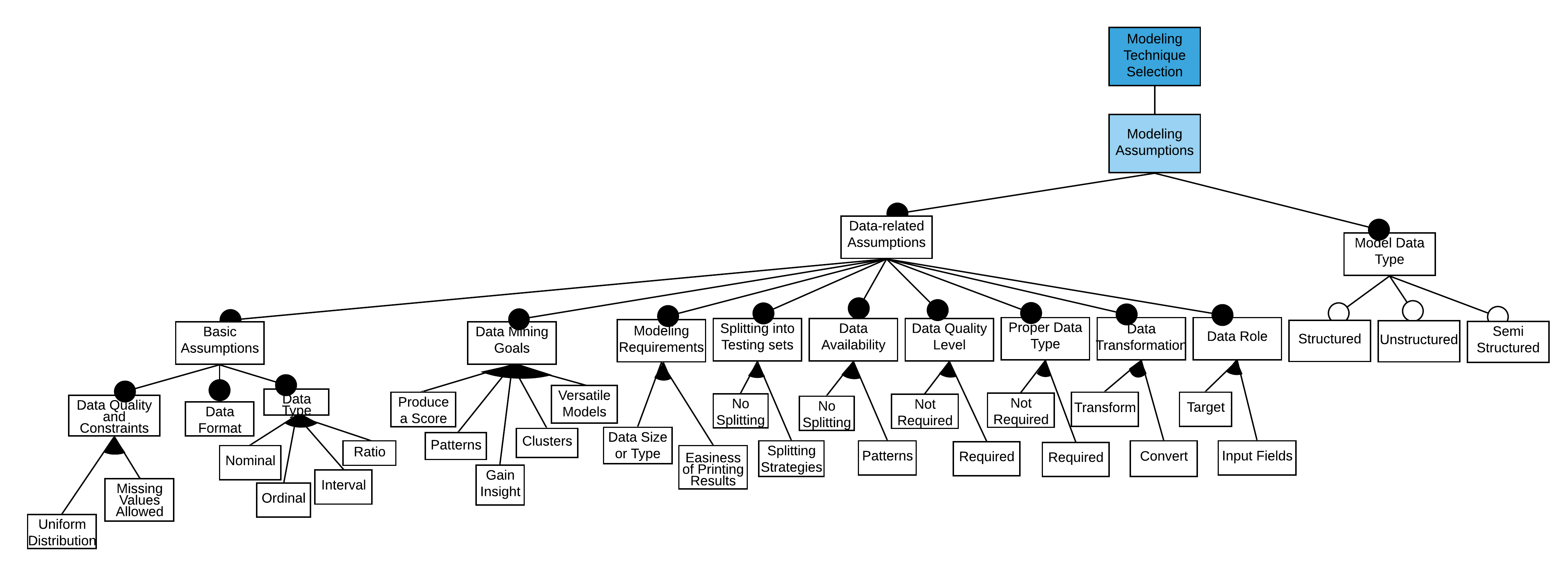}
	\centering
	\caption{Feature diagram for Modeling Technique Selection - Modeling Assumptions.}
	\label{fig:Assumptions}
\end{figure*}

We have modeled the variability of the identified factors using a feature diagram and constraints that trigger adaptive reconfiguration, that is, model selection changes due to variability factors. Figures \ref{fig:ML Techniques} and \ref{fig:Assumptions} illustrate the feature diagrams that we designed to capture various types of variability involved in the process of selecting an ML algorithm. These types of variability include ML algorithm variability (i.e., representing the vast options of algorithms) and modeling assumption variability (i.e., representing data and application variability that can impact the effectiveness of the selected model). The mandatory feature Modeling Assumptions, shown in Fig. \ref{fig:Assumptions}, consists of a group of alternative features which specify assumptions about data according to the modeling technique selected \cite{ncr1999crisp}. Activities for this step involve \cite{ncr1999crisp}: defining any built-in assumptions made by the technique about the data (e.g., quality, format, and distribution) and ensuring that the appropriate model would also need to consider the availability of data types for mining, the data mining goals, and the specific modeling requirements. These elements are optional and might occur depending on the technique selected. For example, applications with the data mining goal of predicting credit risk demand transparent, auditable, and explainable models \cite{ariza2020explainability}.  In this case, decision trees are recommended because of their interpretability  \cite{ariza2020explainability}. In the medical field, applications also require higher levels of safety and explainability. Thus, logistic regression has been encouraged to develop explainable clinical predictive models, even when modern ML models outperform them \cite{zihni2020opening}. 
Based on the notion of optional features, the feature diagram depicts or-group features, namely: Data-related Assumptions, and Model Data Type.

\begin{figure*}[!ht]
	\centering
	\includegraphics[scale=0.13]{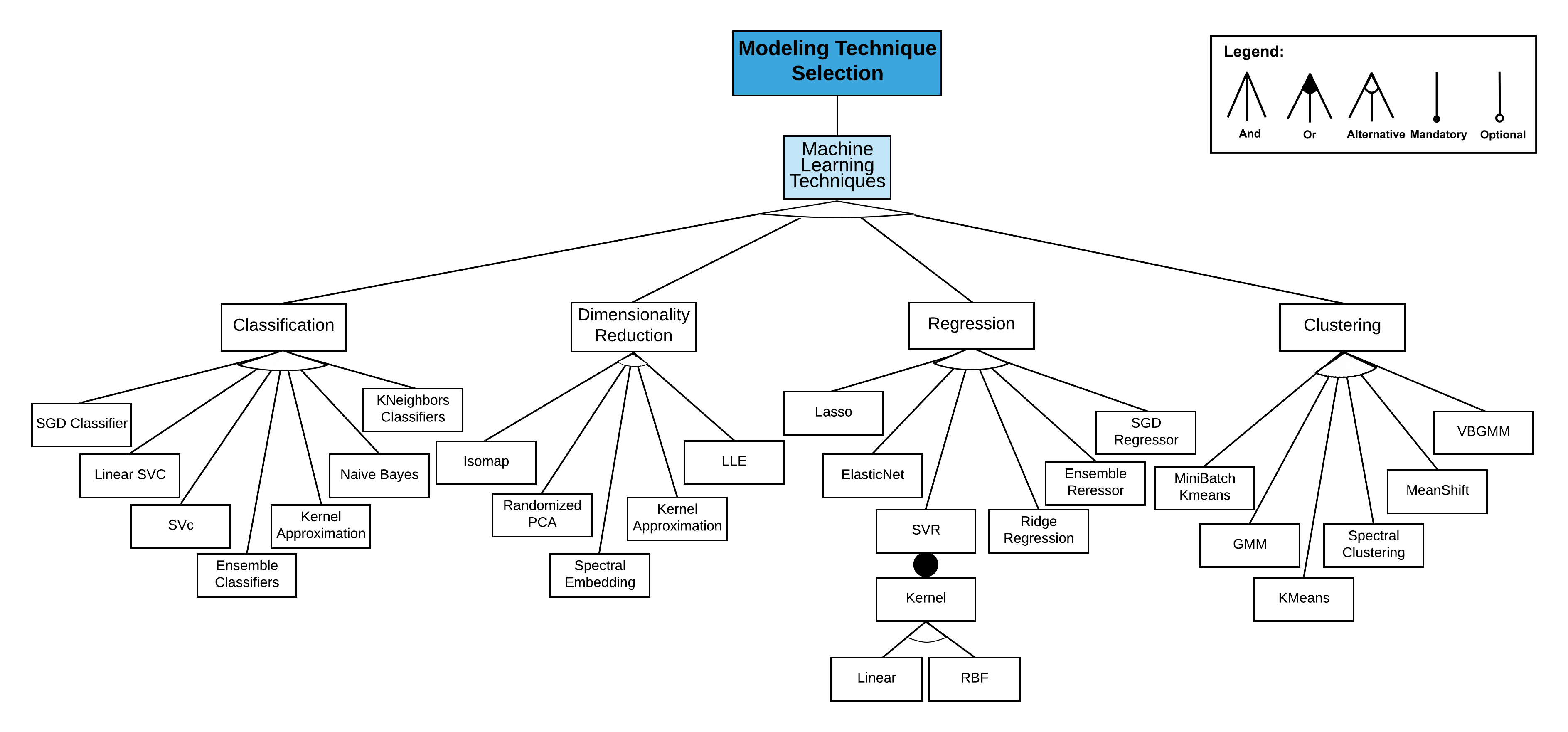}
	\centering
	\caption{Instance feature diagram of the proposed approach to represent Scikit Modeling Techniques feature diagram.}
	\label{fig:scikitinstance1}
\end{figure*}

\begin{figure*}[!ht]
	\centering
	\includegraphics[scale=0.14]{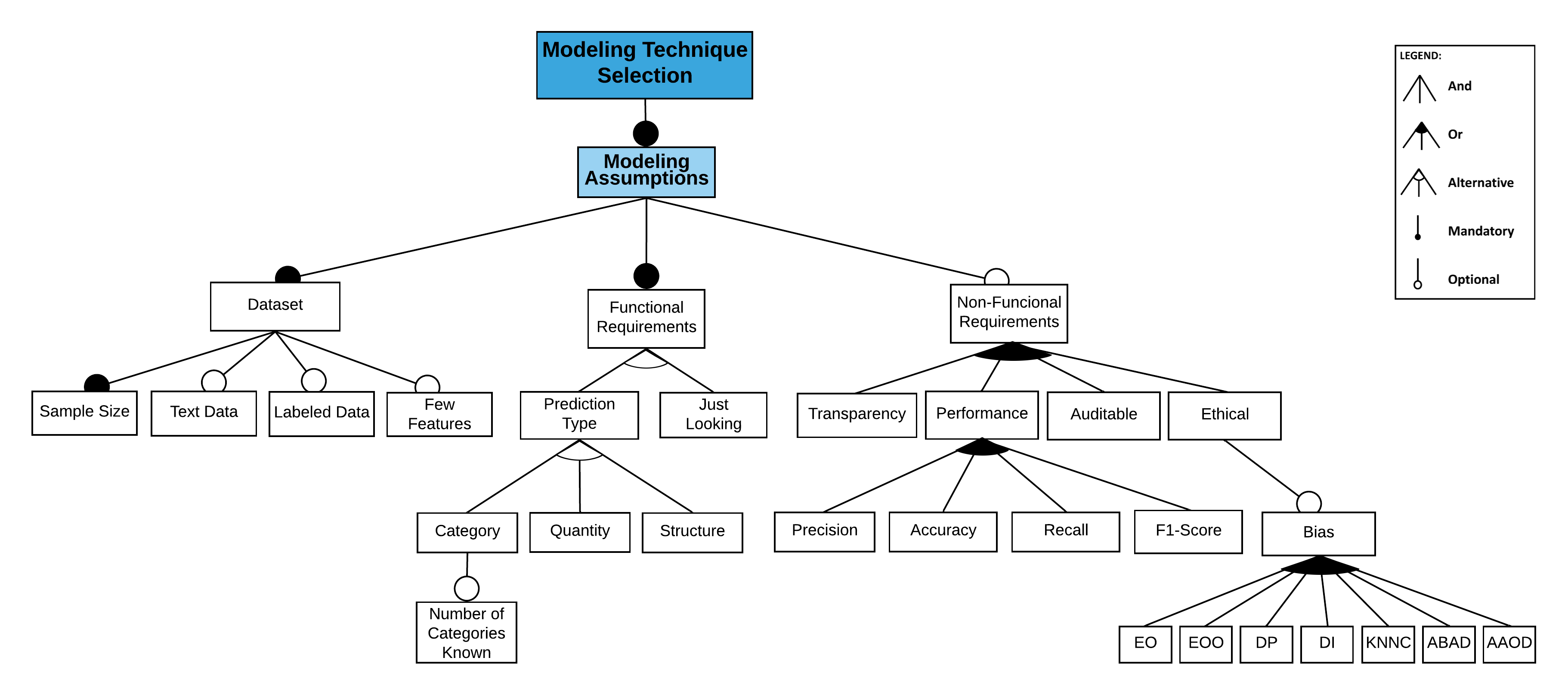}
	\centering
	\caption{Instance feature diagram of the proposed approach to represent Scikit Modeling Assumptions feature diagram.}
	\label{fig:scikitinstance2}
\end{figure*}

Based on the Scikit-Learn heuristics, two feature diagrams were derived. The first diagram, presented in Figure \ref{fig:scikitinstance1}, depicts the feature diagram of the proposed approach to represent Scikit modeling techniques. ML techniques include classification, dimensionality reduction, regression, and clustering.
The second diagram, presented in Figure \ref{fig:scikitinstance2}, shows the feature diagram of the proposed approach to represent the Scikit modeling assumptions. The 'Modeling Assumptions' include dataset requirements (e.g., sample size), functional requirements (e.g., prediction type), and non-functional requirements (e.g., ethical such as bias and its related metrics for detection and mitigation). 
Feature model constraints act as reconfiguration triggers which occur depending on the changes in the application requirements.

\subsection{Model Selection Feature Diagrams with Bias Features} \label{sec:constraints}
In this subsection, we instantiate our variability model with added features related to bias (e.g., bias-related metrics). 
The Ethical feature was added with the bias-related subfeatures that specify possible metrics that detect and mitigate bias and unfairness in ML models. For example, the metric `Equalized Odds' (EO) ensures equal positive results for individual of both positive and negative classes. Another example is `Equality of Opportunity' (EOO), which can be applied to satisfy equal opportunity in a binary classifier. `Disparate Impact' (DI) makes the proportion equal to 1 when comparing the proportion of individuals who receive a favourable outcome for two opposite groups \cite{pagano2022bias}. The other bias metric examples depicted in the diagram refer to K-Nearest Neighbors Consistency (KNNC), Absolute Balanced Accuracy Difference (ABAD), and Absolute Average Odds Difference (AAOD).

\section{Experiments and Results}

To give more details about and illustrate the use of our proposed approach, we selected an example experiment from an existing data science paper by Leenings et al. \cite{leenings2021photonai}. This paper outlines a unique approach for automating the selection of machine learning algorithms and also employs the Scikit-Learn library for testing their solution. By using the same example experiment from Leenings et al. \cite{leenings2021photonai}, we aim to demonstrate how their experiment could conform to the Scikit-Learn feature model, while also allowing us to compare the results. In addition to evaluating performance metrics as done in Leenings et al., we further examine fairness metrics to identify any potential biases or unfairness, drawing upon the methodology outlined in \cite{pagano2022bias}. In healthcare applications, where the reliability of predictions is imperative, addressing model biases becomes essential. Biased models can engender erroneous predictions, compromising the efficacy of healthcare interventions.

\subsection{Dataset: Heart Failure Prediction}

The aim of this application is to forecast the survival outcomes of heart failure patients based on their clinical data, as detailed in Chicco et al. \cite{chicco2020machine}. The dataset provided comprises information from 299 heart failure patients and includes 12 clinical variables, such as age, sex, diabetes status, and serum creatinine levels. These features serve as predictors for mortality due to heart failure. Additionally, the dataset incorporates a target variable named `death\_event,' which is a binary indicator specifying whether the patient survived or died within an average follow-up period of 130 days \cite{chicco2020machine}.

As mentioned in \cite{leenings2021photonai}, this dataset is imbalanced, because the number of survived patients (death event = 0) is 203, while the dead patients (death event = 1) is 96. In statistical terms, there are 32.11\% positives and 67.89\% negatives.

To compare our results with those presented in \cite{leenings2021photonai} and in \cite{chicco2020machine}, we also split the dataset into 80\% (239) for the training set, and 20\% (the remaining 60 patients) for the test set. To split the dataset, we used a stratified split, which is a recommended approach for unbalanced datasets. 

To identify which machine learning algorithms are most suitable for this specific application, the initial step involves instantiating the abstract Dataset class tailored to this application. The corresponding code for this action is shown below.

\begin{lstlisting}[language=Python,style=customc]
class DatasetInstance02(Dataset):  
    def setPredictionType(self):
        self.prediction_type =  `category' 
        self.known_prediction_categories = True
        
    def setDataType(self):
        self.data_type = `numeric'
        
    def setTargetColumn(self):
        self.target_column = `DEATH_EVENT'
        
    def setPandasDataset(self):
        self.dataset = pd.read_csv( `dataset/heart_failure.csv')
        
    def isFewFeatures(self):
        if(len(self.dataset.columns)<30):
            return True
        return False
        
    def setTestDatasetSize(self):
        self.test_dataset_size = 0.2
\end{lstlisting}

It is important to note that we conducted an empirical analysis to establish the threshold for the number of features. In our configuration, we categorize datasets with fewer than 30 features as ``small-feature" datasets.
By implementing this configurable system, we can easily tailor the Dataset class to meet the unique requirements of different applications, offering a flexible and adaptable approach to automated machine learning algorithm selection.


\subsection{Automatic Selection of a Machine Learning Algorithm: Performance Metric}
After setting up the Dataset Class specific to this application, we ran the Method Selector, designed in line with the heuristics outlined in Section \ref{Sec:Approach}. This process yields a FIFO (First-In, First-Out) queue of suggested machine learning algorithms. In this case, the queue consists of LinearSVC, KNeighborsClassifier, SVC, and EnsembleClassifiers, followed by the placeholder `Tough luck.' The system will sequentially test each of these algorithms until it meets a pre-defined quality criteria.

For this specific application, we employed the F1 score as our quality metric, which serves as a measure of the algorithm's performance. The F1 score is particularly important in a medical context, as it balances sensitivity and specificity. Our F1 score target was set at 0.77, based on a baseline of 0.76 established in prior work by Leenings et al \cite{leenings2021photonai}.


\subsubsection{Initial Program Iteration: Output Results}

Our system accurately identified the task at hand as a classification problem and suggested LinearSVC as the most appropriate algorithm for the given dataset. Notably, the performance metrics achieved with LinearSVC exceeded those reported by Leenings et al \cite{leenings2021photonai}. in their study, where their best outcomes were obtained using the Random Forest algorithm with an F1 score of 0.76, Matthews correlation coefficient of 0.662, balanced accuracy of 0.8368, sensitivity of 0.81, and specificity of 0.85.

Performance Metrics of Our LinearSVC Model:
\begin{itemize}
    \item Balanced Accuracy (BACC): 0.848
\item Matthews Correlation Coefficient: 0.672
\item Accuracy: 0.85
\item Sensitivity: 0.854
\item Specificity: 0.842
\item F1 Score: 0.780
\end{itemize}

\subsubsection{Performance metric: Results Discussion}

Leenings et al. \cite{leenings2021photonai} executed the Random Forest classifier, gradient boosting and the support vector classification (SVC) with linear kernel to this dataset in order to identify the one that provides the best F1 score. After executing these algorithms, the authors identified that the Random Forest Classifier presented the best performance. The original paper \cite{chicco2020machine} also proposes the use of Random Forest. However, as Chicco and Jurman \cite{chicco2020machine} do not use an approach to deal with the dataset imbalance, they achieved a lower F1 score of 0.547.


Although the SVC with linear kernel algorithm is similar to the Linear SVC algorithm, which is the one that we used in our tests, they differ in terms of flexibility in the choice of penalties and loss functions \cite{scikit-learn}. 
The two other algorithms tested by Leenings et al. \cite{leenings2021photonai}, Random Forest Classifier and Gradient Boosting, are examples of Ensemble Classifiers. According to the Scikit-Learn flowchart, LinearSVC, KNeighborsClassifier, and SVC are more highly recommended for this dataset than Ensemble Classifiers. 

Leenings et al. \cite{leenings2021photonai} also tested the LassoFeatureSelection for automatically selecting features from the dataset and they observed a decrease on the system performance. According to the Scikit-Learn constraints, the Lasso algorithm is recommended as a regression algorithm. 

\subsection{BIAS Investigation}
While maximizing the F1 score might lead to an algorithm that performs well overall, it does not necessarily guarantee that the algorithm is unbiased. To address this, we introduce fairness metrics and examine how our selected algorithm performs in terms of fairness.

\begin{figure*}[!htb]
\centering
\includegraphics[scale=0.13]{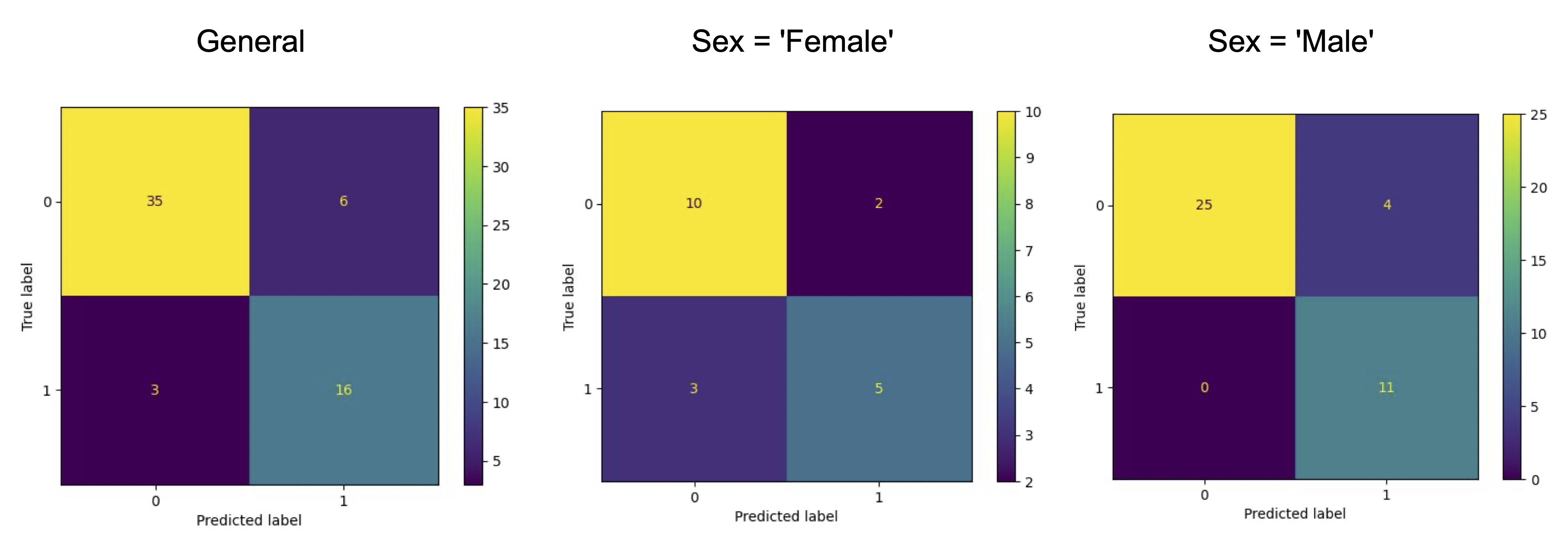}
\centering
\caption{Confusion matrix for extracting parameters for fairness metrics.}
\label{fig:confusion}
\end{figure*}

We divided the validation set into two gender-based groups: men, which we represent as the protected group (P), and women, as the unprotected group (U). The confusion matrix parameters for each group are illustrated in Figure \ref{fig:confusion}, and are crucial for calculating the fairness metrics employed in this study.

\subsubsection{Equality of Opportunity (EOO)}

Equality of Opportunity aims for the True Positive Rate (TPR) to be the same for both protected and unprotected groups. Mathematically, this is expressed as:

\begin{equation}
EOO = \frac{{\text{{tp}}_P}}{{\text{{tp}}_P + \text{{fn}}_P}} - \frac{{\text{{tp}}_U}}{{\text{{tp}}_U + \text{{fn}}_U}}
\end{equation}

For our data, \(EOO = 0.375\). This value indicates a noticeable disparity in True Positive Rates between the protected (P) and unprotected (U) groups. A value of 0 would have indicated perfect fairness in terms of opportunities for both groups to get a favorable outcome. Therefore, the model appears to be biased in terms of gender-based opportunity.

\subsubsection{Disparate Impact (DI)}

Disparate Impact compares the proportion of favorable outcomes in the protected and unprotected groups. For fairness, this ratio should be close to 1:

\begin{equation}
DI = \frac{{\text{{tp}}_P + \text{{fp}}_P}}{{N_P}} \Bigg/ \frac{{\text{{tp}}_U + \text{{fp}}_U}}{{N_U}}
\end{equation}

In our case, \(DI = 1.0714\). The DI value of 1.0714 is close to 1, which suggests that the ratio of favorable outcomes between the protected and unprotected groups is nearly balanced. However, being greater than 1 means that the protected group (men, in this case) slightly has more favorable outcomes than the unprotected group (women).

\subsubsection{Absolute Balanced Accuracy Difference (ABAD)}

The ABAD is the difference in balanced accuracy between the protected and unprotected groups. It is defined as:

\begin{equation}
ABAD = \left(\frac{1}{2} \times (\text{{tp}}_P + \text{{tn}}_P)\right) - (\text{{tp}}_U + \text{{tn}}_U)
\end{equation}

For our dataset, \(ABAD = 3\). The ABAD value of 3 indicates that the balanced accuracy varies between the protected and unprotected groups. The higher the ABAD, the more significant the disparity in balanced accuracy between the two groups. Thus, the value of 3 highlights that there exists a disparity that should be addressed.

\subsubsection{Fairness Evaluation of Queue-Based KNeighborsClassifier Selection}

Upon advancing to the next method in the FIFO queue—selected based on the specified metric—we turn our focus to KNeighborsClassifier. The KNeighborsClassifier shows a mixed performance in fairness metrics compared to the previous algorithm. While its EOO value of 0.1705 suggests improvement in offering equal opportunities between groups, its DI and ABAD scores (0.7 and 4.0, respectively) indicate increased disparity. These contrasting results underline the intricacies involved in balancing multiple fairness dimensions and signal the need for additional model recalibration.

\subsubsection{Fairness Evaluation of Queue-Based SVC Selection}
In the FIFO queue, the next method is SVC, which shows promising fairness metrics with an EOO of 0.0 and a DI of 1.0, suggesting near-ideal fairness. However, its F1 score is quite low at 0.481, emphasizing that while fairness is crucial, effectiveness as measured by traditional metrics cannot be ignored. 

This highlights the need for a composite metric or strategy that harmonizes both social fairness and performance efficacy, ensuring that the selected model excels in both dimensions.

\subsection{Configurable ML Algorithm Selection via Feature Diagrams: Addressing Bias}

\begin{figure*}[!htb]
\centering
\includegraphics[scale=0.047]{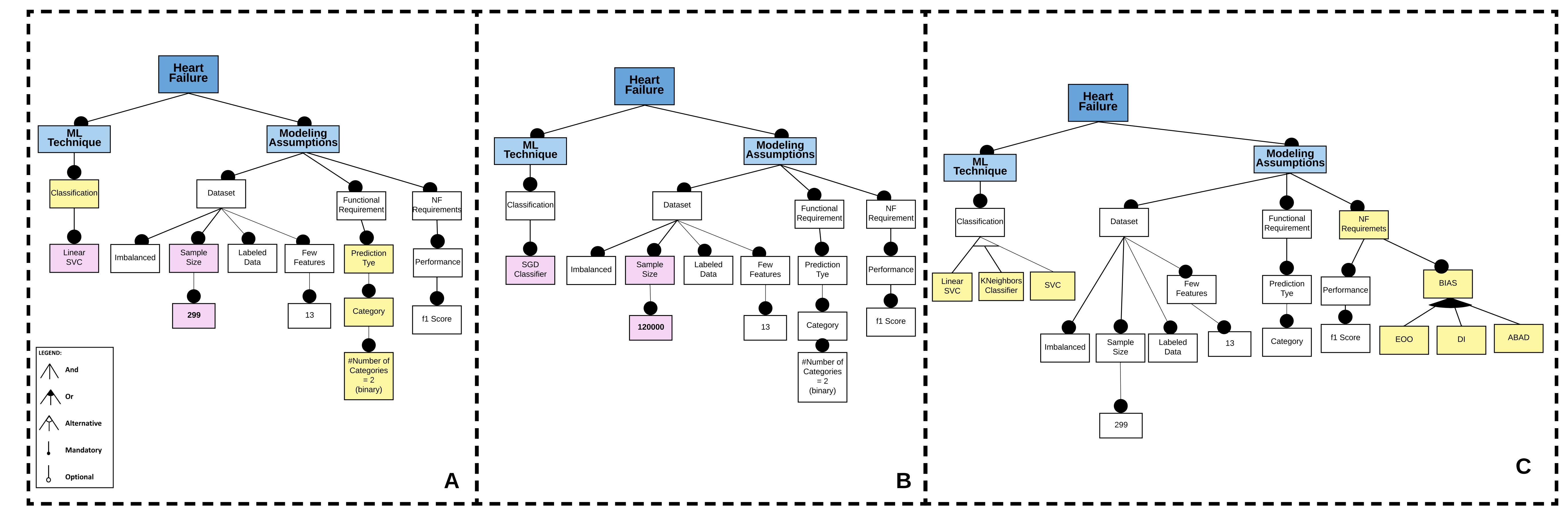}
\caption{Three examples of instances for the Heart Failure Prediction application, highlighting the configuration implemented.}
\label{fig:configurationinstance}
\end{figure*}

Feature diagrams offer a structured and visual framework to capture the complexities inherent in machine learning applications. By utilizing these diagrams, we not only highlight the intricate interplay between modeling assumptions and algorithm choices but also emphasize their implications on both performance and fairness. Balancing these two crucial aspects becomes pivotal, and with the aid of feature diagrams, such a task is made intuitive and effective. Figure \ref{fig:configurationinstance} exemplifies these concerns, presenting configurations influenced by diverse modeling assumptions and the inherent relationship between feature selection, algorithmic decisions, and the biases they might propagate.

\begin{itemize}
    \item Figure A: Based on the F1 Score metric, Linear SVC is chosen for a dataset size of 299.
    \item Figure B: In a scenario with an increased dataset size of 120,000, the SGD classifier is selected due to its efficiency with larger datasets.
    \item Figure C: When fairness metrics like EOO, DI, and ABAD are integrated, the importance of not only considering bias is evident. The F1 Score remains crucial, and the metric choice is dependent on the classification type.
\end{itemize}

Metrics not only influence algorithm selection but also profoundly affect application outcomes. In contrast, outcomes can also dictate the relevance of the metrics chosen. For instance, a surge in data volume can precipitate a change in the chosen algorithm, as exemplified by the transition from Linear SVC to the SGD classifier. Furthermore, changes in evaluation metrics can influence the algorithm selection. The dynamism in this relationship accentuates the need for a flexible model selection framework, particularly in critical domains like healthcare, to concurrently optimize for predictive precision and minimize biases.

\subsubsection{Fairness Metrics: Addressing Variability and Bias}
The analysis of fairness metrics—EOO, DI, and ABAD—offers not only a nuanced view of the model's ethical dimensions but also has profound implications on the inherent variability of machine learning model selection and its assumptions. The observed disparities in EOO, DI, and ABAD may trigger reconfigurations in the feature diagram, influencing the selection or de-selection of specific attributes, thereby affecting the final chosen model and its performance metrics.

To illustrate, an EOO metric valued at 0.375 combined with an ABAD metric of 3 could necessitate re-evaluating dataset equitability and feature prioritization. This might induce a transition from models that prioritize F1 scores to those emphasizing fairness metrics. A DI metric of 1.0714, proximate to a fairness baseline, might lead to reassessing dataset partitioning strategies, influencing model robustness across diverse cross-validation folds.

These recalibrations suggest that fairness metrics are interconnected with other variability factors like feature dimensionality and dataset balance. Altering one quality metric like fairness can cause cascading effects, necessitating system reconfigurations to achieve both high predictive accuracy and ethical fairness. This integrated approach offers a dynamic and adaptive model selection process.

\section{Conclusions and Future Work}
In this paper, we have extended a variability-aware model selection method to cope with bias detection in ML projects.
The method was illustrated by a case study based on heart failure prediction to illustrate how the extended model can be instantiated based on the represented bias features and interactions with other features relevant to model selection.
In general, representing and automating variability-aware ML model selection and bias detection associated with ML projects can benefit both designers and practitioners. Explicit representations cope with the need for transparency and explanability in the model selection processs, and can lead to cost and time savings resulting from reuse and advance the state of the art by constituting a step towards transforming the ML model selection process from an ad hoc  into a more systematic, adaptive and explainable process. 

Although we have focused on bias detection caused by metrics that address the lack of complete and accurate data, many other types of bias can be studies as future work. Other sources of bias can be studied in the context of the variability-aware ML model selection method, including other bias coming from the data (e.g., social), the ML algorithms (e.g., popularity), and the interaction with users (e.g., behaviour). Further, the interactions among the various bias features with other features of the model can be investigated and explicitly represented as feature diagram constraints.
This could lead to more explicit evaluations of the adaptive and explainable aspects of the variability-aware model selection method. It would be interesting to extend the model with bias metrics based on a given context (e.g., pre-processing, processing, and post-processing). 
Finally, the model can be extended in different ways, to be able to represent additional data, algorithm, as well as functional and non-functional requirements such as those related to trust and ethics.   

\section*{Acknowledgment}
This work was supported by the Natural Sciences and Engineering Research Council of Canada (NSERC), and the Centre for Community Mapping (COMAP).

\bibliographystyle{IEEEtran}
\bibliography{sigproc}


\end{document}